\documentclass{article}

\usepackage{microtype}
\usepackage{graphicx}

\usepackage{subcaption}
\usepackage{booktabs} 
\usepackage{hyperref}

\usepackage[preprint]{icml2026}

\usepackage{amsmath}
\usepackage{amssymb}
\usepackage{mathtools}
\usepackage{amsthm}

\usepackage[capitalize,noabbrev]{cleveref}

\theoremstyle{plain}

\theoremstyle{definition}

\theoremstyle{remark}

\usepackage{algorithm}
\usepackage{algorithmic}

\icmltitlerunning{Mechanistic Interpretability of Brain-to-Speech Models Across Speech Modes}

\begin{document}

\twocolumn[
\icmltitle{Mechanistic Interpretability of Brain-to-Speech Models Across Speech Modes}



  \icmlsetsymbol{equal}{*}

  \begin{icmlauthorlist}
    \icmlauthor{Maryam Maghsoudi}{equal,xxx}
    \icmlauthor{Ayushi Mishra}{equal,yyy}
    
  \end{icmlauthorlist}

  \icmlaffiliation{yyy}{Department of Computer Science, University of Maryland, College Park, United States}
  \icmlaffiliation{xxx}{Department of Electrical and Computer Engineering, University of Maryland, College Park, United States}

  \icmlcorrespondingauthor{Maryam Maghsoudi}{maryam00@umd.edu}
  \icmlcorrespondingauthor{Ayushi Mishra}{amishr13@umd.edu}

  \vskip 0.3in
]



%
\printAffiliationsAndNotice{\icmlEqualContribution}

\begin{abstract}
Brain-to-speech decoding models demonstrate robust performance in vocalized, mimed, and imagined speech; yet, the fundamental mechanisms via which these models capture and transmit information across different speech modalities are less explored. In this work, we use mechanistic interpretability to causally investigate the internal representations of a neural speech decoder. We perform cross-mode activation patching of internal activations across speech modes, and use tri-modal interpolation to examine whether speech representations vary discretely or continuously. We use coarse-to-fine causal tracing and causal scrubbing to find localized causal structure, allowing us to find internal subspaces that are sufficient for cross-mode transfer. In order to determine how finely distributed these effects are within layers, we perform neuron-level activation patching. We discover that small but not distributed subsets of neurons, rather than isolated units, affect the cross-mode transfer. Our results show that speech modes lie on a shared continuous causal manifold, and cross-mode transfer is mediated by compact, layer-specific subspaces rather than diffuse activity. Together, our findings give a causal explanation for how speech modality information is organized and used in brain-to-speech decoding models, revealing hierarchical and direction-dependent representational structure across speech modes.

\end{abstract}

\section{Introduction}

Brain–computer interfaces (BCIs) that decode speech directly from neural activity hold the promise of restoring natural communication to individuals who are unable to speak. Recent advances in neural decoding models have demonstrated impressive performance in reconstructing acoustic speech features from neural recordings, particularly for overt or vocalized speech \cite{neural_decoding1}\cite{neural_decoding2}\cite{neural_decoding3}\cite{neural_decoding4}\cite{neural_decoding5}. However, practical and clinical deployment of speech BCIs ultimately depends on decoding speech in settings where vocalization is absent or limited, such as silently mimed or purely imagined speech which is challenging due to variability in timing alignment \cite{cnn}.

Despite growing empirical success, a fundamental gap remains in our understanding of how neural speech decoding models internally represent and process different speech modes. Models trained on vocalized, mimed, and imagined speech exhibit large performance differences, but it is unclear whether these differences arise from irreducible properties of the neural inputs themselves or from how internal representations are formed and utilized within the model. Without a mechanistic account of these representations, progress in transfer learning, robustness, and generalization across speech modes remains largely empirical and ad hoc.

The machine learning community has seen rapid progress in mechanistic interpretability, a framework for understanding neural networks through causal interventions on internal activations rather than post hoc correlations \cite {mi}. Techniques such as activation patching, causal tracing, and subspace interventions have enabled precise attribution of model behavior to internal representations in domains such as language modeling and vision \cite {activation_patching1, activation_patching2}. However, these tools have rarely been applied to brain-to-speech decoding models \cite{open_problem}, where understanding internal mechanisms is arguably as important as improving raw performance.

This study uses mechanistic interpretability methods and use controlled interventions to find out how the network represents and transfers information about each speech mode. The neural speech decoder is trained on three types of speech: vocalized, mimed, and imagined. By moving internal activations between modes while keeping the inputs the same, we test which representations are needed for accurate decoding. The study also looks at whether these representations change in clear steps or gradually across speech modes, and whether important information is spread out or focused in certain parts of the network.

Our results show that differences between speech modes are encoded along shared, graded internal representations. High-performing vocalized representations can be causally transferred to imagined inputs to recover decoding quality, while the reverse intervention catastrophically degrades performance. We show that cross-mode transfer is mediated by compact and structured internal subspaces, both temporally in recurrent dynamics and spatially in convolutional features, rather than diffuse network-wide activity.

This work moves beyond descriptive performance comparisons and toward principled understanding. Our findings have implications not only for improving imagined-speech BCIs, but also for designing models that are interpretable, robust, and capable of reliable cross-mode generalization.

\section{Related Work}

Recent progress in explainable artificial intelligence (XAI) has made it possible to use deep learning systems and the large language model (LLM) in many clinical decision-making settings \cite{llm1}\cite{llm2}. In these areas, models need to interpret complex biomedical signals in a reliable and transparent way \cite{bci1}\cite{bci2}\cite{bci3}\cite{bci4}\cite{bci4}\cite{bci5}. It is important to understand how these models make their predictions to ensure safety and trust \cite{bci6}\cite{bci7}\cite{bci8}. Speech brain–computer interfaces (BCIs) are an example of this challenge, as they use neural networks to decode speech information from brain recordings. DeWave \cite{dewave} explores speech and language decoding from EEG, including EEG-to-text translation and marker-free decoding, but primarily emphasizes end-task performance. These approaches do not examine how cross-mode performance differences arise within the trained decoder or whether they are driven by limitations of the input signals or by internal model representations.

Mechanistic interpretability tries to reverse-engineer learned representations and computations by changing internal activations, instead of just using correlational probes \cite{mi5}. Activation patching, also known as causal tracing or interchange intervention, is a key method here \cite{activation_patching3}. Recent best-practice advice highlights the importance of choosing baselines, corruption distributions, and metrics carefully \cite{mi1}\cite{mi2}\cite{mi3}\cite{mi4}. Circuit-oriented methods go further than single-layer interventions by finding causal components and their interactions, often using automated workflows for circuit discovery and attribution-based localization \cite{sparse}. Several surveys and position papers bring these methods together and discuss open problems, such as the need for clearer concepts and more reliable causal interventions. In this work, we use these causal tools in a neural decoding setting and focus on cross-condition transfer between speech modes as a natural way to intervene.

Recent research suggests that interpretability should be more closely connected to cognitive and computational neuroscience \cite{neuroai}, where understanding internal representations is a main scientific goal rather than a secondary concern. In BCI studies, researchers often analyze electrode localization, feature importance, or architectural ablations \cite{audio_models}. These approaches can show where signals are found, but they usually cannot tell us if certain internal representations are actually needed or enough for decoding. By using activation-level interventions, causal localization, and scrubbing, we provide a clearer explanation of how speech-mode information is represented in a trained decoder. Our approach complements existing performance-focused studies by revealing how internal model representations, rather than input signals alone, govern decoding behavior across speech modes.

\section{Mechanistic Interpretability across Speech Modes}

Neural speech decoding models increasingly achieve strong performance across diverse speech conditions, including vocalized, mimed, and imagined speech [\textbf{Appendix~\ref{baseline}}]. However, it remains unclear \emph{how} such models internally represent different speech modes.



Our goal is not merely to measure correlations between internal activations and output behavior, but to identify which internal representations are \emph{causally responsible} for speech decoding across modes. To this end, we employ intervention-based analyses inspired by recent work on activation patching and causal tracing in mechanistic interpretability \cite{mi1}. These methods allow us to modify specific internal activations while holding the input and remaining computation fixed, enabling counterfactual reasoning about the role of internal representations.

Crucially, the presence of multiple speech modes provides a natural setting for causal analysis. Because vocalized, mimed, and imagined speech share high-level linguistic content but differ in neural signal properties and decoding performance, they allow us to ask whether internal representations learned under one mode can substitute for, degrade, or interpolate with those from another. This cross-mode structure enables interventions that go beyond standard ablations, directly testing sufficiency, necessity, and continuity of internal representations across conditions.


\section{Dataset and Model Architecture}

\subsection{A Stereotactic EEG Dataset for Vocalized, Mimed, and Imagined Speech}

We evaluate our methods on \textsc{VocalMind} \cite{He2025VocalMind}, a large-scale stereotactic EEG (sEEG) dataset designed for neural speech decoding across multiple speech modes. The dataset contains neural recordings from a human participant performing speech tasks under three conditions: \emph{vocalized}, \emph{mimed}, and \emph{imagined} speech.
\textbf{Vocalized speech} (overt speech) involves spoken articulation with sound production.
\textbf{Mimed speech} consists of silent articulation without acoustic output.
\textbf{Imagined speech} involves internal speech generation without overt articulation; here we focus on speech motor imagery corresponding to imagined speaking.

Unlike most prior neural speech datasets that focus on a single speech mode, \textsc{VocalMind} enables systematic comparison across related speaking conditions. This multi-modal structure is well-suited for studying representational transfer and causal mechanisms across speech modes.



\subsection{Model Architecture}

\begin{figure}[t]
    \centering
    \includegraphics[width=\linewidth]{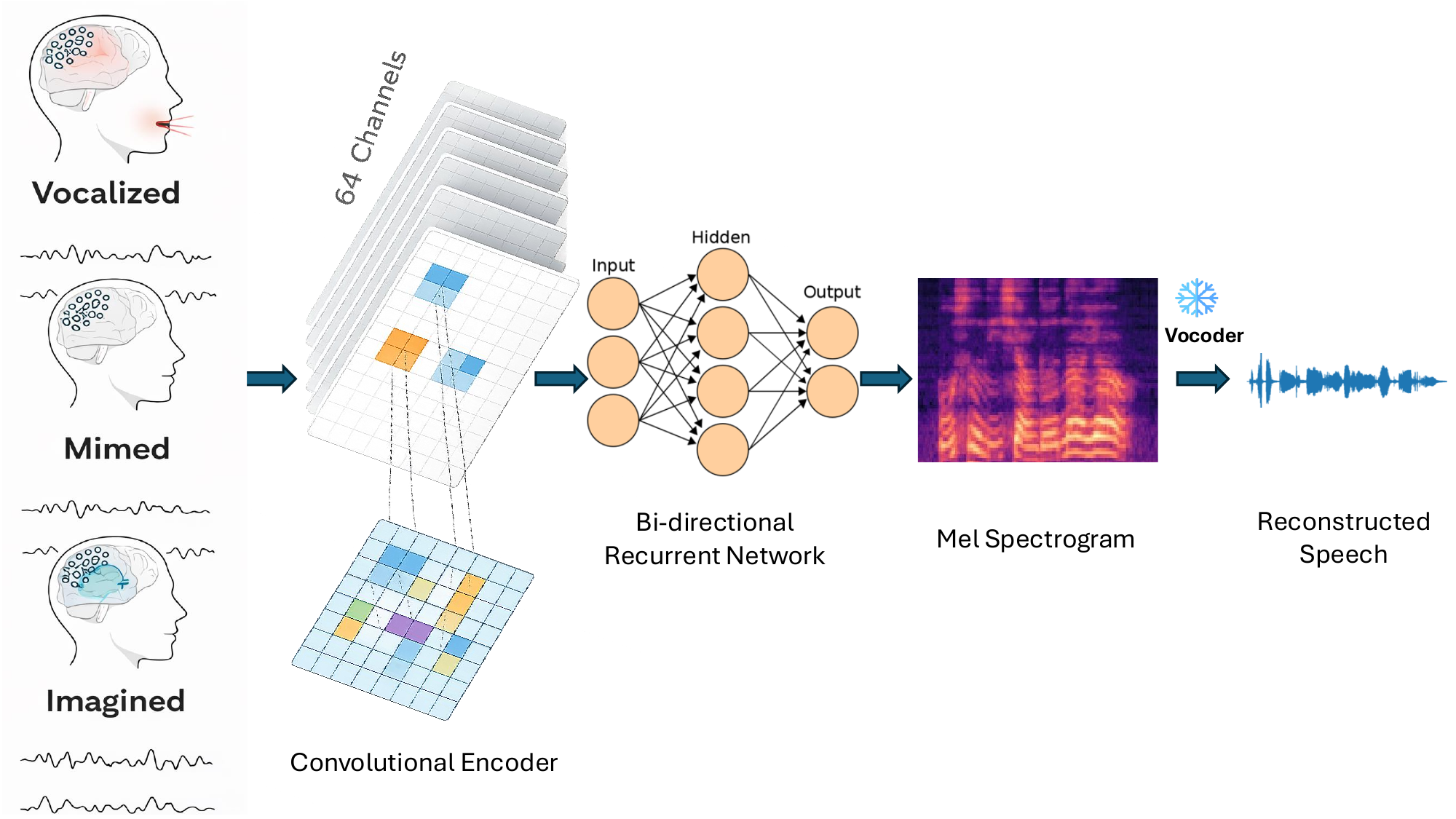}
    \caption{Multi-channel sEEG signals are processed by a convolutional encoder to extract spatiotemporal features, followed by a bidirectional recurrent network that integrates information over time and predicts frame-level mel-spectrograms. A neural vocoder converts predicted spectrograms into reconstructed speech.}
    \label{fig:architecture}
\end{figure}

The model consists of three main components as shown in Figure~\ref{fig:architecture}. First, a convolutional encoder extracts local spatiotemporal features from raw sEEG signals across channels and time. Concretely, this encoder begins with a one-dimensional convolutional layer (\texttt{Conv1D}) operating along the temporal dimension, with 64 output channels, a kernel size of 4, a stride of 4, and padding of 2, thereby performing temporal downsampling while preserving local temporal structure. 
The resulting convolutional activations are then passed to a multi-layer bidirectional recurrent network, which integrates information over longer temporal contexts and produces frame-level acoustic predictions.

The recurrent component consists of a gated recurrent unit (GRU) network with 64 input features and 256 hidden units per direction, arranged in three stacked layers. 
Finally, a fully connected linear layer maps the recurrent features to an 80-dimensional output space corresponding to mel-spectrogram frames. A neural vocoder \cite{vocoder} subsequently reconstructs the speech waveform from the predicted mel-spectrogram.

Let $f(x)$ denote the full model, where $x$ is an sEEG input sequence. We denote by $z^{\text{conv}}$ the activations produced by the convolutional encoder (\texttt{conv\_out}) and by $z^{\text{rnn}}$ the activations produced by the recurrent network (\texttt{rnn\_out}). These internal representations form the primary intervention sites for the mechanistic analyses described in the following sections. Unless otherwise stated, the recurrent component is implemented as a multi-layer bidirectional GRU. 

\vspace{-1mm}
\section{Cross-Mode Activation Patching}

We study the causal role of internal representations across speech modes using \emph{cross-mode activation patching}. Let $X_I$ denotes imagined and $X_V$ denotes vocalized speech mode. For a given layer $\ell$, let $Z_I^{\ell}$ and $Z_V^{\ell}$ denote the corresponding hidden activations produced by a standard forward pass. 

\begin{figure*}[t]
    \centering
    \includegraphics[width=\textwidth]{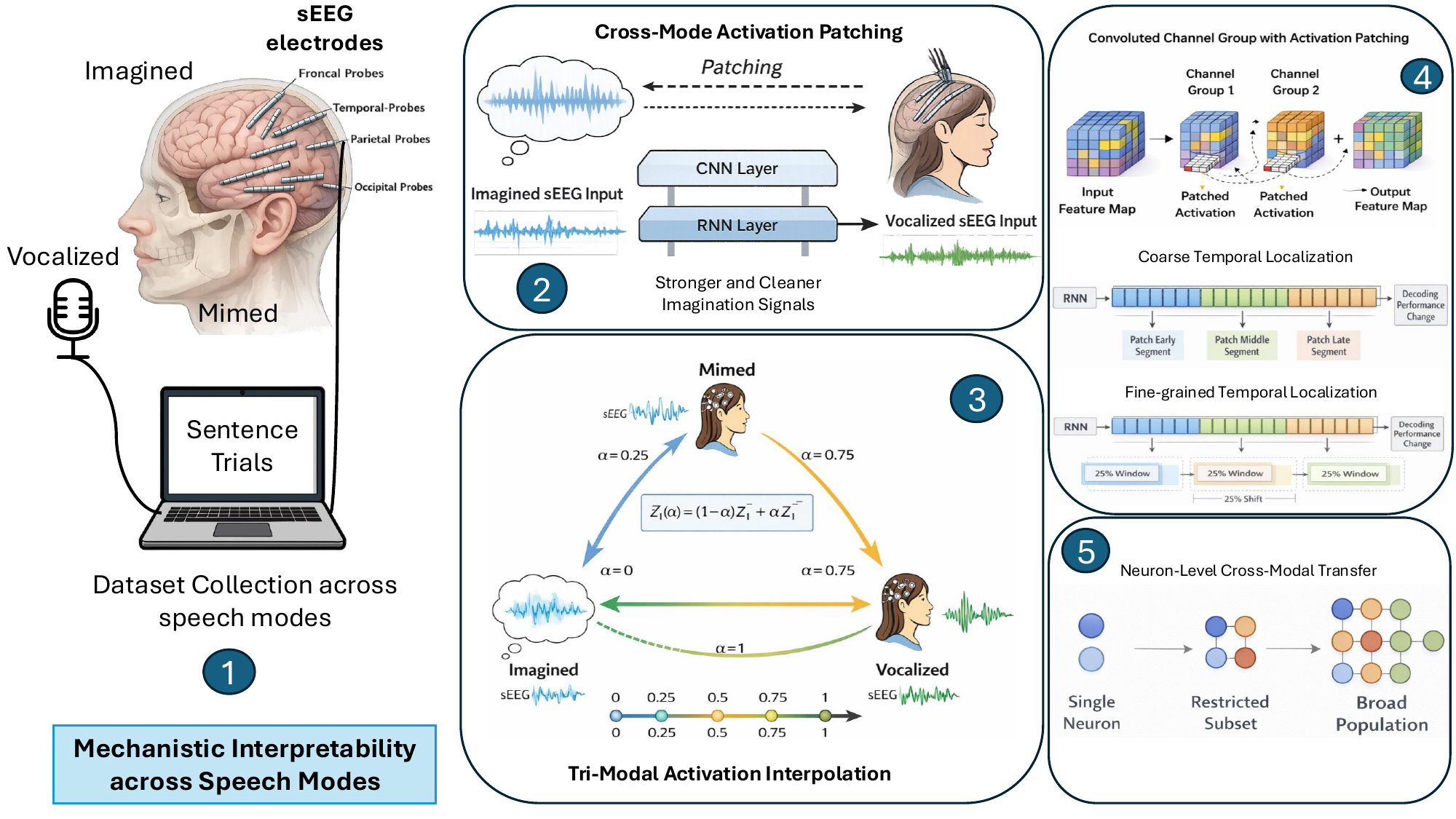}
    \caption{\textbf{Mechanistic interpretability across speech modes.}
    \textbf{(1)} We study speech decoding under three modes: \emph{vocalized}, \emph{mimed}, and \emph{imagined}, recorded with sEEG.
    \textbf{(2)} \emph{Cross-mode activation patching:} we replace internal activations from one mode with those from another at selected layers (e.g., CNN/RNN) to test whether representations transfer across modes and whether patching yields stronger or cleaner imagination signals.
    \textbf{(3)} \emph{Tri-modal activation interpolation:} we linearly interpolate activations between modes to probe smooth transitions in representational geometry and decoding behavior.
    \textbf{(4)} \emph{Temporal localization:} we causally localize when information matters by patching subsequent convolution channels and contiguous early/middle/late thirds of the RNN activation sequence (coarse), and by sliding a 25\% window with 25\% shift (fine-grained), measuring the resulting change in decoding performance.
    \textbf{(5)} \emph{Neuron-level cross-modal transfer:} we test whether transfer is mediated by single neurons, restricted subsets, or broad populations by patching individual neurons and then patching top-$k$ neurons grouped by their effects.}
    \label{fig:overview_speech_modes}
\end{figure*}

In a patched forward pass, we replace the activation $Z_I^{\ell}$ with $Z_V^{\ell}$ while keeping the original input $X_I$ fixed, and propagate the modified activation through the remainder of the network. Let $f_{\text{post-}\ell}$ denote the subnetwork consisting of all layers following layer $\ell$. The resulting patched output is given by
\vspace{-1mm}
\begin{equation}
\tilde{y} = f_{\text{post-}\ell}(z_v^{\ell}).
\end{equation}
We apply this procedure to both convolutional and recurrent layers to test two complementary causal properties: \emph{sufficiency} and \emph{necessity}. Sufficiency is evaluated by patching activations from a higher-performing speech mode (e.g., vocalized speech) into a lower-performing input (e.g., imagined speech), and measuring whether decoding performance improves. Necessity is evaluated by patching activations from a lower-performing mode into a higher-performing input and assessing whether performance degrades.

This bidirectional cross-mode patching framework enables direct causal attribution of performance differences to internal representations at specific network layers. Our full mechanistic story is described in Figure ~\ref{fig:overview_speech_modes}.


\subsection{Tri-Modal Activation Interpolation}

Cross-mode activation patching can test whether causal sufficiency and necessity are discrete, but it does not show if internal representations change gradually across speech modes. To find out if speech modes are encoded as separate categories or along a continuum, we use tri-modal activation interpolation with Imagined (I), Mimed (M), and Vocalized (V) speech.

For a given sentence shared across modes, we first run standard forward passes to extract internal activations at a target layer $\ell \in \{\text{conv\_out}, \text{rnn\_out}\}$, yielding representations $z_\ell^A$ and $z_\ell^B$ from two speech modes $A$ and $B$. We then construct interpolated activations as a convex combination:
\begin{equation}
\tilde{z}_\ell(\alpha) = (1 - \alpha) z_\ell^A + \alpha z_\ell^B, \quad \alpha \in [0,1].
\end{equation}

During a patched forward pass, the interpolated activation $\tilde{z}_\ell(\alpha)$ replaces the original activation at layer $\ell$, while the input EEG signal is fixed to mode $A$. Decoding performance is evaluated at multiple interpolation coefficients $\alpha \in \{0, 0.25, 0.5, 0.75, 1\}$.

We perform this procedure for all ordered pairs among Imagined, Mimed, and Vocalized speech. 

This intervention directly tests whether internal representations across speech modes lie on a shared causal manifold, and whether differences between modes reflect graded variation or qualitatively distinct internal states.

To further understand where within the network cross-modal information is instantiated, we next perform a more detailed causal tracing analysis.

\subsection{Causal Localization and Fine-Grained Tracing}

Building on cross-mode activation patching and interpolation, we next localize where speech-relevant causal information is instantiated within the network. We perform both coarse interventions to identify candidate regions and fine-grained tracing to resolve their internal structure.

\paragraph{Coarse causal localization.}
At the convolutional layer, which produces 64 feature channels, we partition channels into four groups of 16 and patch one group at a time while holding the remainder fixed. This tests whether causal signal is distributed across channels or concentrated within specific feature subspaces.  
At the recurrent layer, we divide the \texttt{rnn\_out} sequence into early, middle, and late temporal segments and patch each segment independently, probing whether causal information is localized to a particular phase of recurrent processing.

\begin{figure}[t]
    \centering
    \includegraphics[width=1\linewidth]{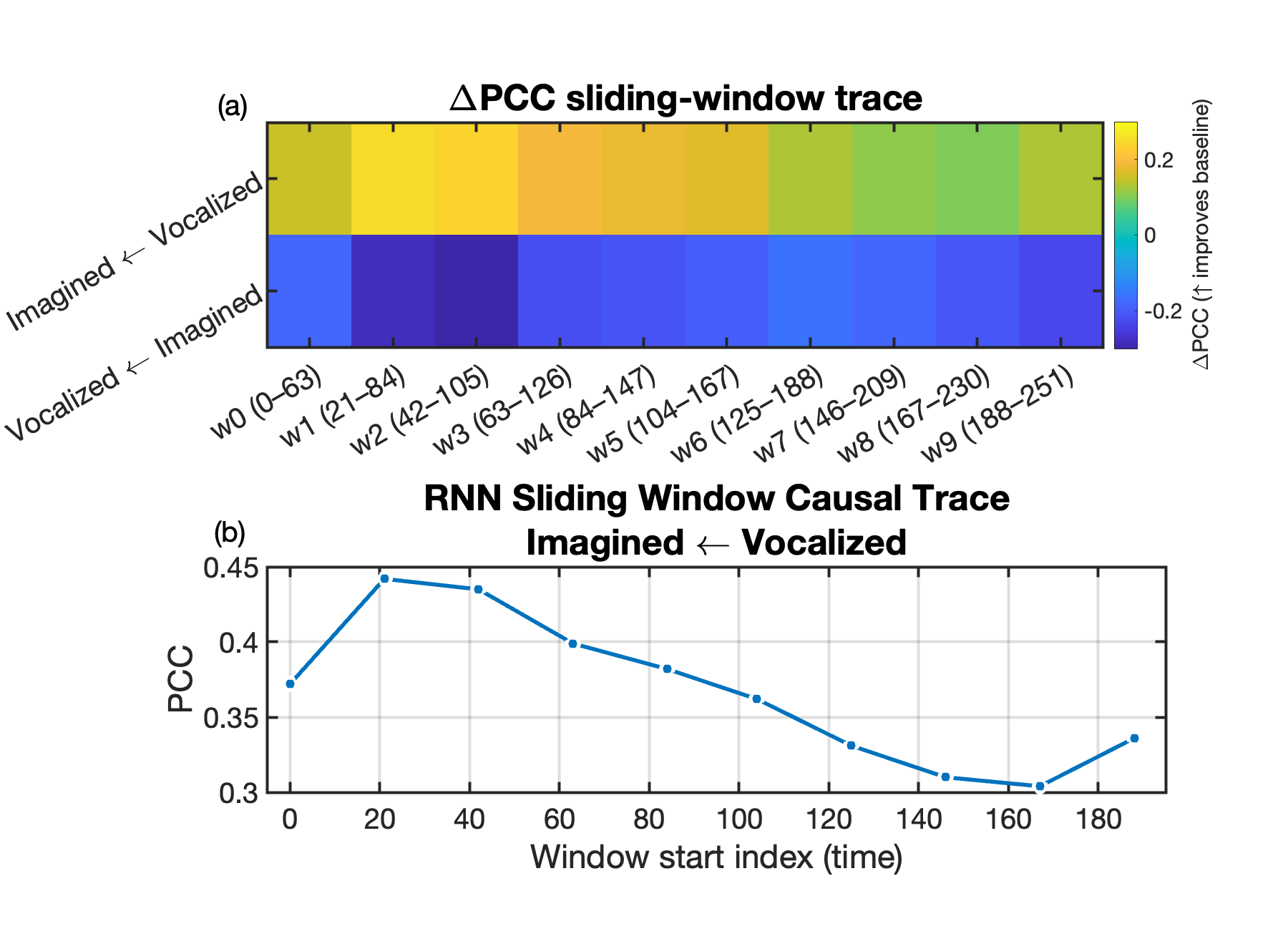}
    \caption{
    RNN sliding-window causal tracing across speech modes. Early-to-mid recurrent states form a causal bottleneck that is both sufficient and necessary for high-quality speech decoding.}
    \label{fig:rnn_trace}
\end{figure}

\paragraph{Fine-grained causal tracing.}
We refine these analyses using targeted interventions. For the RNN, we apply a sliding-window causal trace by patching a 25\% temporal window and shifting it across 10 positions along the 256-dimensional sequence. This produces a causal profile over time. As shown in Figure~\ref{fig:rnn_trace}, transferring vocalized RNN states to imagined inputs yields the strongest gains for early-to-mid windows, specifically for window time steps from [21-84], while later windows have weaker effects; the necessity direction exhibits a complementary degradation pattern in the same region.

\begin{figure}[t]
    \centering
    \includegraphics[width=0.95\linewidth]{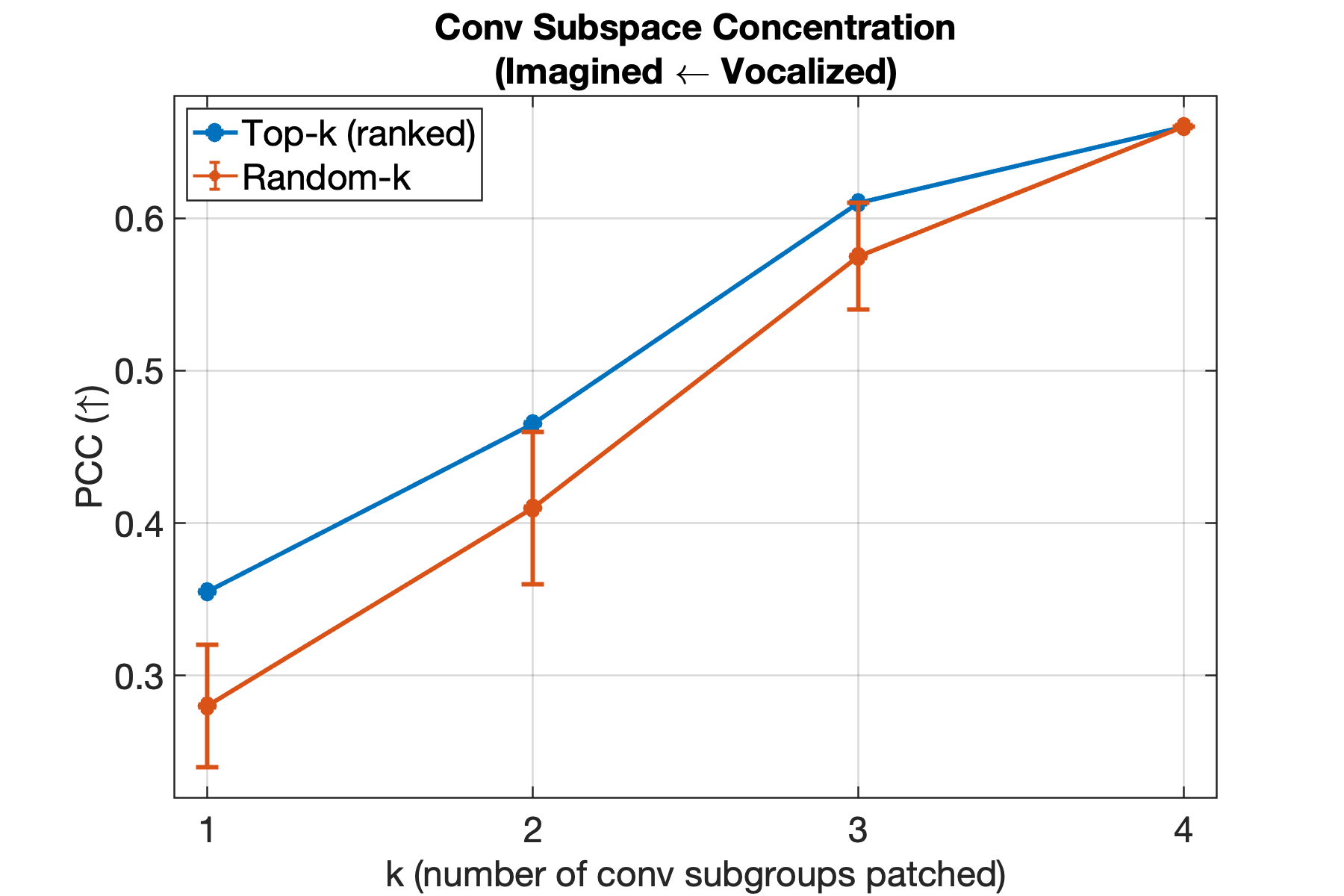}
    \caption{
    Causal subspace concentration in the convolutional encoder.
    Ranked Top-$k$ channel subgroup patching consistently outperforms matched random-$k$ controls, demonstrating that cross-mode transfer is mediated by a compact convolutional subspace rather than uniformly distributed features.
    }
    \label{fig:topk_random}
\end{figure}

We test if the causal influence in the convolutional layer is focused in a low-dimensional subspace. After finding the most effective coarse channel group (g2; channels 32 to 48) [\textbf{Appendinx~\ref{causal_loc}}], we split it into smaller groups and rank them by patching benefit. Figure~\ref{fig:topk_random} shows that cumulative Top-$k$ patching leads to steady improvements and always does better than matched random-$k$ controls. This suggests that cross-mode transfer happens through a compact convolutional subspace instead of spread-out features.

\label{fine-grained}



\subsection{Causal Scrubbing for Localized Subspaces}

We use \emph{causal scrubbing} to test whether the internal subspaces identified by activation patching and fine-grained causal tracing (Section~\ref{fine-grained}) are sufficient for cross-mode transfer, and whether information outside these subspaces is replaceable. For each stimulus key shared across modes, we first compute baseline decoding for Vocalized, Mimed, and Imagined speech. We then select a donor mode and intervene while decoding a different recipient mode by patching internal activations at \texttt{conv\_out} (channels) and \texttt{rnn\_out} (time).

Based on coarse-to-fine causal tracing, we define small keep regions at each site: (i) convolutional channels $[32,48)$ (16 channels), and (ii) recurrent time steps $[21,84)$ (25\% of the sequence).

For each recipient input, a hybrid activation is made by copying donor activations in the keep region and filling the rest with activations from another randomly chosen donor example of the same shape. The three versions tested are \textbf{KEEP-Conv}, \textbf{KEEP-RNN}, and \textbf{KEEP-Combo}. Each one keeps either the chosen convolutional subspace, the recurrent time window, or both.


To control for the amount of patched information instead of subspace identity, we use matched random controls (\textbf{RAND-Conv}, \textbf{RAND-RNN}, \textbf{RAND-Combo}) that keep a randomly chosen block of the same size. We also include full-patch baselines (\textbf{fullConv}, \textbf{fullRNN}), where the entire \texttt{conv\_out} or \texttt{rnn\_out} activation is replaced with donor activations. This gives an upper limit on possible transfer.

\subsection{Neuron-level Activation Patching}
\label{sec:neuron_patching}
The investigations up to this point were conducted at the coarse level of network layers. In this section, we move to a finer scale and focus on smaller units within layers - individual neurons. We ask two related questions.

First, is cross-modal transfer mediated by a single neuron, a restricted subset of neurons, or by broad population-level patching? To address this question, we perform single-neuron activation patching across modes and measure the resulting changes in decoding performance caused by patching each neuron individually. We then extend this analysis by grouping neurons based on their individual effects and patching the top-$k$ neurons together, to test whether improvements or degradation arise from a small subset of neurons acting jointly.

Second, are the most influential neurons consistent across different input samples? To answer this question, we identify the most influential neuron for each individual sample and analyze the distribution and reuse of these neurons across sentences.

We first describe the activation patching setup and then present two experiments addressing the questions above.

\paragraph{Setup.}
Consider a source input mode $m_s$ (patch-from), a target input mode $m_t$ (patch-to), a layer $\ell$, and a neuron index $i$. We write the decoder as a composition of modules and denote by $a_\ell(x)$ the activation tensor produced by module $\ell$ when processing an sEEG input $x$. Consider paired inputs $(x^{(m_s)}, x^{(m_t)}, y)$ that share the same ground-truth target $y$ (mel-spectrogram). The source and target activations at layer $\ell$ are
$A_s = a_\ell\left(x^{(m_s)}\right) \in \mathbb{R}^{T_\ell \times D_\ell}$ and
$A_t = a_\ell\left(x^{(m_t)}\right)$, respectively. Where $T_\ell$ denotes the number of time steps at layer $\ell$ and $D_\ell$ is the dimensionality of the layer’s hidden representation (i.e., the number of neurons)

We define a patched activation $\tilde A_t^{(i)}$ by replacing only the $i$-th neuron across all time steps:
\begin{equation} \tilde A_t^{(i)}[t, j] = \begin{cases} A_s[t, i] & \text{if } j=i,\\ A_t[t, j] & \text{if } j\neq i, \end{cases} \qquad \forall t \in \{1,\dots,T_\ell\}. \end{equation}
We run the target-mode input through the \emph{same} fixed model while intervening at layer $\ell$ to substitute neuron $i$ with the source activation trajectory. The resulting patched prediction is
\begin{equation}
\hat y^{(i)} =
M_{\theta}\left(x^{(m_t)};\ a_\ell(x^{(m_t)}) \leftarrow \tilde A_t^{(i)}\right).
\end{equation}
For comparison, we also compute the unpatched baseline prediction
$\hat y^{\mathrm{base}} = M_{\theta}\left(x^{(m_t)}\right)$,
and quantify improvements or degradation relative to $\hat y^{\mathrm{base}}$.
This procedure isolates the causal contribution of a single neuron’s temporal activation trajectory to cross-modal transfer.

\subsection{Top-$k$ neuron saturation analysis}
\label{sec:saturation}
Single-neuron patching reveals the effect of individual neurons on decoding performance. However, to assess whether performance changes are driven by coordinated subsets of neurons rather than isolated units, we perform a top-$k$ neuron saturation analysis. In this analysis, for a given layer $\ell$ and modality pair $(m_s \rightarrow m_t)$, we first rank neurons based on their mean single-neuron effect across the dataset, $\overline{\Delta \mathrm{PCC}}_{i}$, producing an ordering $\pi$ from most to least causally influential.

We then define the top-$k$ neuron set $S_{k}={\pi(1),\dots,\pi(k)}$, which contains the $k$ neurons whose individual patching yields the largest improvement in decoding performance. We jointly patch all neurons in $S_{k}$ by substituting their activation trajectories from the source pass into the target-model forward pass, while keeping all model parameters fixed. The resulting patched activation at layer $\ell$ is
\begin{equation}
\tilde A_{t}^{(S_k)}[t,j]=
\begin{cases}
A_s[t,j] & j \in S_k,\\
A_t[t,j] & \text{otherwise.}
\end{cases}
\end{equation}
\noindent for all $t \in \{1,\dots,T_\ell\}$.

We evaluate decoding performance as a function of $k$, yielding saturation curves whose shape reflects how causal information is distributed across neurons within a layer.


\subsection{Consistency of neuron-level effects across sentences}
To answer the second question, we assess whether the same neurons are responsible for improvements across different input samples. To do so, we perform a sentence-level winner analysis.
We fix a source–target modality pair $(m_s \rightarrow m_t)$ and a layer $\ell$, and for each input sample we patch all single neurons from the source modality into the target model. For each neuron and each input sample, we measure $\Delta\mathrm{PCC}$ relative to the unpatched baseline. We then select, for each input sample, the \emph{winner neuron} as the neuron that yields the largest improvement in $\Delta\mathrm{PCC}$. This results in one winning neuron index per sentence.

We then analyze the distribution of winning neurons across input samples.
(i) We compute the number of unique winner neurons. If this number is small, it suggests that only a few neurons are responsible for improvements across all input samples; if it is large, it indicates that responsibility is spread across many neurons.
(ii) We compute the number of input samples affected by the most frequently occurring winner neuron.
(iii) We compute how many input samples are covered by the top-$k$ most frequently occurring winner neurons. If these top-$k$ neurons account for a large portion of the samples, it suggests that the same neurons tend to drive improvements across the dataset; otherwise, the neurons responsible for improvements differ from sentence to sentence.
Finally, (iv) we compute the entropy of the winner-neuron distribution. Higher entropy means that neurons are more evenly dispersed and involved in specific sentences.
\begin{table*}[t]
\centering
\begin{tabular}{lcccccc}
\toprule
\textbf{Direction} 
& \textbf{Full Patch} 
& \textbf{KEEP-Conv} 
& \textbf{RAND-Conv} 
& \textbf{KEEP-RNN} 
& \textbf{RAND-RNN} \\
\midrule
\multicolumn{6}{l}{\textit{PCC}} \\
Imagined $\leftarrow$ Vocalized (Suff.) 
& \textbf{0.954} & \textbf{0.666} & 0.564 & \textbf{0.462} & 0.398 \\
Vocalized $\leftarrow$ Imagined (Nec.) 
& \textbf{0.177} & 0.575 & 0.575 & 0.456 & 0.460 \\
\midrule
\multicolumn{6}{l}{\textit{MCD}} \\
Imagined $\leftarrow$ Vocalized (Suff.) 
& \textbf{1.63} & \textbf{3.13} & 3.23 & 3.10 & 3.26 \\
Vocalized $\leftarrow$ Imagined (Nec.) 
& \textbf{3.70} & 2.95 & 2.98 & \textbf{2.88} & 3.15 \\
\bottomrule
\end{tabular}
\caption{Causal scrubbing results for directional interventions only. We compare full activation patching with subspace-preserving scrubbing (KEEP) and matched random controls (RAND). Results are reported for sufficiency (Imagined $\leftarrow$ Vocalized) and necessity (Vocalized $\leftarrow$ Imagined).}
\label{tab:causal_scrubbing_horizontal_directional}
\end{table*}
Together, these measures indicate whether cross-modal improvements are caused by a consistent set of neurons or by a diverse, input-specific subset.

\section{Evaluation}

\paragraph{Metrics.}
We evaluate the causal effect by comparing patched predictions to unpatched baselines using two complementary metrics.
We use the Pearson correlation coefficient (PCC) between the predicted and ground-truth mel-spectrograms to measure global reconstruction fidelity, and Mel Cepstral Distortion (MCD) to measure spectral accuracy.
PCC is computed as the correlation between flattened spectrograms, while MCD is computed from cepstral coefficients obtained by applying a log transform followed by a discrete cosine transform along frequency.
For both metrics, we report changes relative to the baseline prediction, with higher $\Delta$PCC and lower MCD indicating improved performance.

\subsection{A Shared Causal Manifold Across Speech Modes}

Across speech modes, causal interventions reveal a shared internal structure rather than discrete representations. Full cross-mode activation patching shows that vocalized internal states are sufficient to recover high-quality decoding on imagined and mimed inputs, while replacing vocalized states with imagined ones causes performance to collapse. The baseline results for cross-mode activation patching are illustrated in \textbf{Appendix~\ref{baseline2}}. Tri-modal activation interpolation further demonstrates smooth, monotonic changes in decoding quality across imagined, mimed, and vocalized speech, with consistent ordering and symmetry across directions. The performance of all three speech modes varies smoothly with the interpolation coefficient
$\alpha$. The tri-modal interpolation experiments results are discussed in \textbf{Appendix~\ref{tri-modal}}. Together, these results indicate that speech modes lie on a shared continuous causal manifold, with mimed speech occupying an intermediate representational position rather than forming a separate regime.

\subsection{Causal Scrubbing Confirms Concentrated and Directional Transfer}

Causal scrubbing tests whether the subspaces identified by activation patching are sufficient for cross-mode transfer. For Imagined $\leftarrow$ Vocalized, preserving the identified convolutional subspace significantly outperforms matched random controls, with additional but weaker gains from preserving the RNN time window, indicating that transfer is concentrated in a small set of internal features. Table~\ref{tab:causal_scrubbing_horizontal_directional} summarizes the results of the causal scrubbing analysis. In the reverse direction, full patching causes a collapse to imagined-level performance, while subspace-preserving scrubbing yields limited and metric-specific benefits, reflecting an inherent asymmetry in representational completeness between vocalized and imagined speech.

\begin{figure*}[t]
    \centering
    \includegraphics[width=\linewidth]{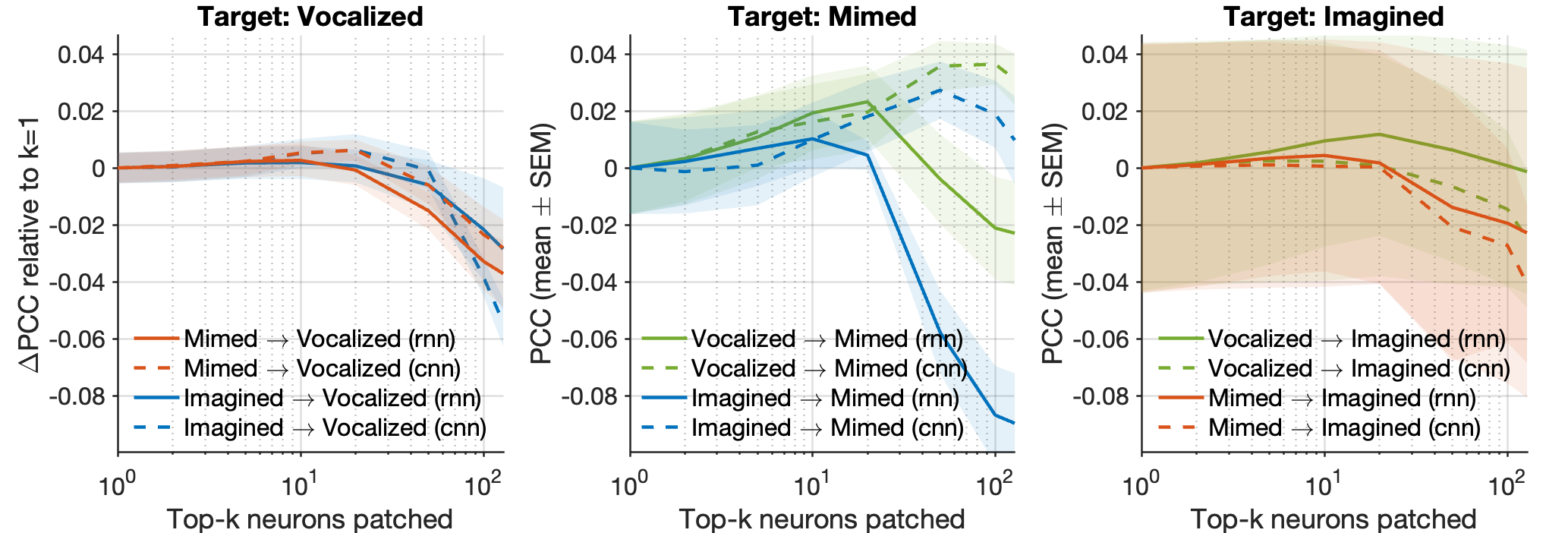}
    \caption{Top-k neuron patching reveals asymmetric subspace transfer across speech modalities.
$\Delta$PCC (relative to k = 1) is shown as a function of the number of patched neurons for three target modalities: Vocalized (left), Mimed (middle), and Imagined (right). Solid lines denote RNN layers and dashed lines denote convolutional layers. Shaded regions indicate $\pm$ SEM across folds.}
    \label{fig:sat_curve}
\end{figure*}

\subsection{Neuron-level Saturation Analysis}
We observe that patching individual neurons does not lead to large changes in PCC, indicating that there are no single neurons whose activation alone drives cross-modal performance. Motivated by this observation, we extend the analysis to patching groups of neurons. Specifically, neurons are first patched individually, ranked according to their single-neuron effect on PCC, and then patched jointly in a top-$k$ manner.

Figure~\ref{fig:sat_curve} shows the change in PCC as a function of $k$. Each subpanel corresponds to a target modality (patched-to), with curves showing patching from the other two source modalities (patched-from). Across all panels, we observe a common pattern of early improvement in $\Delta$PCC followed by degradation as $k$ increases. This saturation–degradation profile indicates that cross-modal transfer does not arise from wholesale replacement of internal representations, but instead depends on a restricted causal subspace whose indiscriminate expansion introduces interference.

The magnitude and direction of these effects depend strongly on the source–target modality pairing. Patching from vocalized activations into mimed or imagined targets yields the largest PCC improvements (green curves in Figure~\ref{fig:sat_curve}), whereas patching into vocalized targets produces little or no gain, suggesting that the decoder is already near its optimal regime for vocalized speech; additional modality-specific asymmetries are analyzed in \textbf{Appendix~\ref{app:modality_asymmetry}}.

Finally, architectural differences appear across conditions. In cases where patching convolutional-layer neurons improves PCC (notably when the target is mimed or imagined), peak performance occurs at larger values of $k$ than in the RNN layers. This delayed peak suggests that convolutional layers support more distributed causal contributions, whereas recurrent layers rely on narrower but more precise subsets of neurons.

\subsection{Sentence-level structure of neuron-level causal effects}
To assess how neuron-level causal effects generalize across sentences, we analyze sentence-level winner coverage as a function of the number of patched neurons ($k$). Coverage increases with $k$ across all source–target modality pairs, indicating that PCC improvements arise from multiple neurons rather than a single dominant unit. Detailed coverage curves and quantitative summaries are provided in \textbf{Appendix~\ref{app:coverage_appendix}}

\vspace{-2mm}
\section{Discussion and Future Work}
Our results show that the performance gap between vocalized, mimed, and imagined speech is mainly due to how the decoder organizes information, not differences in input signal quality. Interpolation shows that the three speech modes are on a shared continuum, with mimed speech always in the middle. Causal interventions show that only vocalized internal states allow for high-quality decoding of imagined inputs and are needed to keep vocalized-level performance. Causal localization finds that cross-mode transfer happens in small bottlenecks, like a limited convolutional subspace and a short part of the recurrent trajectory.

Our neuron-level causal analysis shows that cross-modal speech decoding relies on shared but unevenly distributed neural subspaces. Moreover, vocalized representations provide a more effective subspace for improving mimed and imagined decoding at the neuron level, consistent with the coarse activation-patching results. Future research can link neuron-level causal structure to anatomical and functional organization in sEEG may provide deeper insight into the neural basis of imagined and covert speech.

\paragraph{Future Work.}
Future studies should check if the identified convolutional and recurrent bottlenecks remain when using other sequence models, like LSTMs, transformers, or state-space models, and with more subjects and datasets. Watching how the identified subspaces develop during training could also help explain how model structure affects learning. Training should also encourage imagined and mimed inputs to use the same internal subspaces which support accurate vocalized decoding. This could make the models easier to interpret and help them generalize better across different speech modes.

\vspace{-2mm}
\section{Conclusion}

This work demonstrates that mechanistic interpretability can uncover the internal representations that causally support speech decoding and clarify how these representations differ across vocalized, mimed, and imagined speech. By grounding performance differences in localized and transferable internal structure, our findings point toward neural decoding models that are not only more accurate, but also more predictable, controllable, and robust.

\vspace{-2mm}
\section*{Impact Statement}

This research improves our understanding of how brain-to-speech decoding works by using causal interventions to find out which internal representations affect performance in vocalized, mimed, and imagined speech. These findings can help create more reliable assistive communication systems by supporting thoughtful cross-mode generalization instead of relying on trial and error. However, because neural data is very sensitive, any advances in decoding must include strong privacy measures, informed consent, and clear communication to avoid exaggerating what the models reveal about human cognition.

\newpage
\bibliography{example_paper}

@article{mi,
  title={An empirical comparison of deep learning explainability approaches for EEG using simulated ground truth},
  author={Sujatha Ravindran, Akshay and Contreras-Vidal, Jose},
  journal={Scientific Reports},
  volume={13},
  number={1},
  pages={17709},
  year={2023},
  publisher={Nature Publishing Group UK London}
}

@article{activation_patching1,
  title={Localizing model behavior with path patching},
  author={Goldowsky-Dill, Nicholas and MacLeod, Chris and Sato, Lucas and Arora, Aryaman},
  journal={arXiv preprint arXiv:2304.05969},
  year={2023}
}

@article{activation_patching2,
  title={Towards best practices of activation patching in language models: Metrics and methods},
  author={Zhang, Fred and Nanda, Neel},
  journal={arXiv preprint arXiv:2309.16042},
  year={2023}
}

@article{neural_decoding1,
  title={Neural decoding of EEG signals with machine learning: a systematic review},
  author={Saeidi, Maham and Karwowski, Waldemar and Farahani, Farzad V and Fiok, Krzysztof and Taiar, Redha and Hancock, Peter A and Al-Juaid, Awad},
  journal={Brain sciences},
  volume={11},
  number={11},
  pages={1525},
  year={2021},
  publisher={MDPI}
}

@article{neural_decoding2,
  title={Deep learning approaches for neural decoding across architectures and recording modalities},
  author={Livezey, Jesse A and Glaser, Joshua I},
  journal={Briefings in bioinformatics},
  volume={22},
  number={2},
  pages={1577--1591},
  year={2021},
  publisher={Oxford University Press}
}

@article{neural_decoding3,
  title={Speech decoding from stereo-electroencephalography (sEEG) signals using advanced deep learning methods},
  author={Wu, Xiaolong and Wellington, Scott and Fu, Zhichun and Zhang, Dingguo},
  journal={Journal of Neural Engineering},
  volume={21},
  number={3},
  pages={036055},
  year={2024},
  publisher={IOP Publishing}
}

@article{neural_decoding4,
  title={Decoding semantics from natural speech using human intracranial EEG},
  author={Pescatore, Camille RC and Zhang, Haoyu and Hadjinicolaou, Alex E and Paulk, Angelique C and Rolston, John D and Richardson, R Mark and Williams, Ziv M and Cai, Jing and Cash, Sydney S},
  journal={bioRxiv},
  year={2025}
}

@article{neural_decoding5,
  title={BrainStratify: Coarse-to-Fine Disentanglement of Intracranial Neural Dynamics},
  author={Zheng, Hui and Wang, Hai-Teng and Jing, Yi-Tao and Lin, Pei-Yang and Zhao, Han-Qing and Chen, Wei and Wei, Peng-Hu and Shan, Yong-Zhi and Zhao, Guo-Guang and Liu, Yun-Zhe},
  journal={arXiv preprint arXiv:2505.20480},
  year={2025}
}

@article{He2025VocalMind,
  title = {VocalMind: A Stereotactic EEG Dataset for Vocalized, Mimed, and Imagined Speech in Tonal Language},
  author = {He, Tianyu and Wei, Mingyi and Wang, Ruicong and Wang, Renzhi and Du, Shiwei and Cai, Siqi and Tao, Wei and Li, Haizhou},
  journal = {Scientific Data},
  volume = {12},
  number = {1},
  pages = {657},
  year = {2025},
  publisher = {Nature Publishing Group},
  doi = {10.1038/s41597-025-04741-2},
  pmid = {40253415},
}

@article{bci1,
  title={Interpretable embeddings of speech enhance and explain brain encoding performance of audio models},
  author={Shimizu, Riki and Antonello, Richard J and Singh, Chandan and Mesgarani, Nima},
  journal={arXiv preprint arXiv:2507.16080},
  year={2025}
}

@inproceedings{bci2,
  title={Explainability for Speech Models: On the Challenges of Acoustic Feature Selection},
  author={Fucci, Dennis and Savoldi, Beatrice and Gaido, Marco and Negri, Matteo and Cettolo, Mauro and Bentivogli, Luisa},
  booktitle={Proceedings of the 10th Italian Conference on Computational Linguistics (CLiC-it 2024)},
  pages={373--381},
  year={2024}
}

@article{bci3,
  title={Explainable artificial intelligence techniques for speech emotion recognition: A focus on xai models},
  author={Norval, Michael and Wang, Zenghui},
  journal={Inteligencia Artificial},
  volume={28},
  number={76},
  pages={85--123},
  year={2025}
}

@article{bci4,
  title={Enhancing brain disease diagnosis with XAI: a review of recent studies},
  author={Bibi, Nighat and Courtney, Jane and McGuinness, Kevin},
  journal={ACM Transactions on Computing for Healthcare},
  volume={6},
  number={2},
  pages={1--35},
  year={2025},
  publisher={ACM New York, NY}
}

@inproceedings{bci5,
  title={Explainable artificial intelligence on biosignals for clinical decision support},
  author={Maurer, Miriam Cindy and Metsch, Jacqueline Michelle and Hempel, Philip and Bender, Theresa and Spicher, Nicolai and Hauschild, Anne-Christin},
  booktitle={Proceedings of the 30th ACM SIGKDD Conference on Knowledge Discovery and Data Mining},
  pages={6597--6604},
  year={2024}
}

@article{bci6,
  title={Interpretable multi-timescale models for predicting fmri responses to continuous natural speech. bioRxiv},
  author={Jain, Shailee and Vo, Vy A and Mahto, Shivangi and LeBel, Amanda and Turek, Javier S and Huth, Alexander G},
  year={2021}
}

@article{bci7,
  title={The cognitive revolution in interpretability: From explaining behavior to interpreting representations and algorithms},
  author={Davies, Adam and Khakzar, Ashkan},
  journal={arXiv preprint arXiv:2408.05859},
  year={2024}
}

@article{bci8,
  title={From Black Box to Biomarker: Sparse Autoencoders for Interpreting Speech Models of Parkinson's Disease},
  author={Plantinga, Peter and Chen, Jen-Kai and Sattari, Roozbeh and Ravanelli, Mirco and Klein, Denise},
  journal={arXiv preprint arXiv:2507.16836},
  year={2025}
}

@article{dewave,
  title={Dewave: Discrete eeg waves encoding for brain dynamics to text translation},
  author={Duan, Yiqun and Zhou, Jinzhao and Wang, Zhen and Wang, Yu-Kai and Lin, Chin-Teng},
  journal={arXiv preprint arXiv:2309.14030},
  year={2023}
}

@article{cnn,
  title={A Convolutional Framework for Mapping Imagined Auditory MEG into Listened Brain Responses},
  author={Maghsoudi, Maryam and Rezaeizadeh, Mohsen and Shamma, Shihab},
  journal={arXiv preprint arXiv:2512.03458},
  year={2025}
}

@article{llm1,
  title={EEG-GPT: exploring capabilities of large language models for EEG classification and interpretation},
  author={Kim, Jonathan W and Alaa, Ahmed and Bernardo, Danilo},
  journal={arXiv preprint arXiv:2401.18006},
  year={2024}
}

@inproceedings{llm2,
  title={SING: Spatial Context in Large Language Model for Next-Gen Wearables},
  author={Mishra, Ayushi and Bai, Yang and Narayanasamy, Priyadarshan and Garg, Nakul and Roy, Nirupam},
  booktitle={Forty-second International Conference on Machine Learning}
}

@article{activation_patching3,
  title={How to use and interpret activation patching},
  author={Heimersheim, Stefan and Nanda, Neel},
  journal={arXiv preprint arXiv:2404.15255},
  year={2024}
}

@article{mi1,
  title={Mechanistic interpretability for ai safety--a review, 2024},
  author={Bereska, Leonard and Gavves, Efstratios},
  journal={URL https://arxiv. org/abs/2404.14082}
}

@article{mi2,
  title={Towards automated circuit discovery for mechanistic interpretability},
  author={Conmy, Arthur and Mavor-Parker, Augustine and Lynch, Aengus and Heimersheim, Stefan and Garriga-Alonso, Adri{\`a}},
  journal={Advances in Neural Information Processing Systems},
  volume={36},
  pages={16318--16352},
  year={2023}
}

@article{mi3,
  title={Beyond Transcription: Mechanistic Interpretability in ASR},
  author={Glazer, Neta and Segal-Feldman, Yael and Segev, Hilit and Shamsian, Aviv and Buchnick, Asaf and Hetz, Gill and Fetaya, Ethan and Keshet, Joseph and Navon, Aviv},
  journal={arXiv preprint arXiv:2508.15882},
  year={2025}
}

@article{sparse,
  title={Sparse autoencoders find highly interpretable features in language models},
  author={Cunningham, Hoagy and Ewart, Aidan and Riggs, Logan and Huben, Robert and Sharkey, Lee},
  journal={arXiv preprint arXiv:2309.08600},
  year={2023}
}

@article{mi4,
  title={Mechanistic Interpretability Needs Philosophy},
  author={Williams, Iwan and Oldenburg, Ninell and Dhar, Ruchira and Hatherley, Joshua and Fierro, Constanza and Rajcic, Nina and Schiller, Sandrine R and Stamatiou, Filippos and S{\o}gaard, Anders},
  journal={arXiv preprint arXiv:2506.18852},
  year={2025}
}

@article{neuroai,
  title={NeuroAI for AI safety},
  author={Mineault, Patrick and Zanichelli, Niccol{\`o} and Peng, Joanne Zichen and Arkhipov, Anton and Bingham, Eli and Jara-Ettinger, Julian and Mackevicius, Emily and Marblestone, Adam and Mattar, Marcelo and Payne, Andrew and others},
  journal={arXiv preprint arXiv:2411.18526},
  year={2024}
}

@article{mi5,
  title={Unboxing the black box: Mechanistic interpretability for algorithmic understanding of neural networks},
  author={Kowalska, Bianka and Kwa{\'s}nicka, Halina},
  journal={arXiv preprint arXiv:2511.19265},
  year={2025}
}

@article{audio_models,
  title={Interpretable embeddings of speech enhance and explain brain encoding performance of audio models},
  author={Shimizu, Riki and Antonello, Richard J and Singh, Chandan and Mesgarani, Nima},
  journal={arXiv preprint arXiv:2507.16080},
  year={2025}
}

@article{vocoder,
  title={Hifi-gan: Generative adversarial networks for efficient and high fidelity speech synthesis},
  author={Kong, Jungil and Kim, Jaehyeon and Bae, Jaekyoung},
  journal={Advances in neural information processing systems},
  volume={33},
  pages={17022--17033},
  year={2020}
}

@article{open_problem,
  title={Open problems in mechanistic interpretability},
  author={Sharkey, Lee and Chughtai, Bilal and Batson, Joshua and Lindsey, Jack and Wu, Jeff and Bushnaq, Lucius and Goldowsky-Dill, Nicholas and Heimersheim, Stefan and Ortega, Alejandro and Bloom, Joseph and others},
  journal={arXiv preprint arXiv:2501.16496},
  year={2025}
}
\bibliographystyle{icml2026}

\newpage
\appendix
\onecolumn
\clearpage
\section{Neural speech decoder performance across Vocalized, Imagined, and Mimed speech modes}
\label{baseline}

\begin{figure*}[h]
    \centering
    \includegraphics[width=0.85\textwidth]{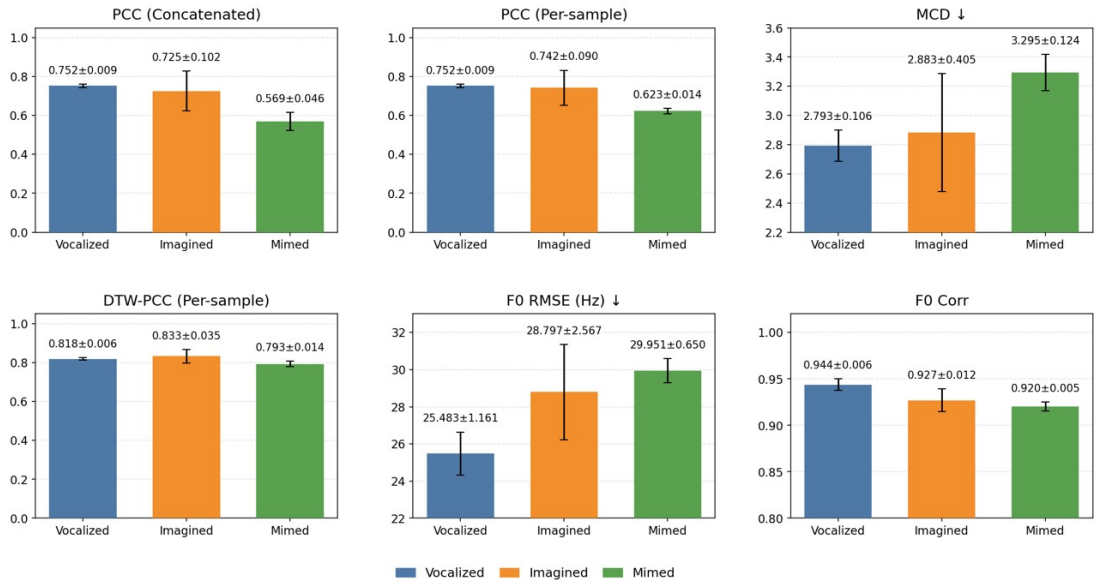}
    \caption{Baseline decoding performance across speech modes (mean $\pm$ std). Higher is better for PCC/DTW-PCC/F0 Corr; lower is better for MCD/F0 RMSE.}
    \label{fig:baseline_metrics}
\end{figure*}

Across modes, the baseline decoder shows clear performance differences consistent across evaluation metrics (Figure.~\ref{fig:baseline_metrics}). Vocalized speech achieves the strongest reconstruction quality, with the highest PCC and DTW-PCC and the lowest spectral distortion (MCD) and pitch error (F0 RMSE). Imagined speech degrades relative to vocalized, while remaining comparable in some correlation-based measures, and exhibits higher variability across samples. Mimed speech is consistently the most challenging condition overall, with the lowest PCC/DTW-PCC and the highest MCD and F0 RMSE. These baselines motivate our causal interventions, which test whether cross-mode gaps are attributable to internal representations rather than input differences.

Table~\ref{tab:baseline_metrics} reports baseline reconstruction quality across speech modes (mean $\pm$ std). Vocalized achieves the best overall performance, imagined is intermediate with higher variance, and mimed is consistently the most challenging condition.

\begin{table}[h]
\centering
\small
\setlength{\tabcolsep}{6pt}
\begin{tabular}{lccc}
\toprule
\textbf{Metric} & \textbf{Vocalized} & \textbf{Imagined} & \textbf{Mimed} \\
\midrule
PCC (concat) $\uparrow$ & $0.7519 \pm 0.0089$ & $0.7254 \pm 0.1021$ & $0.5686 \pm 0.0459$ \\
PCC (per-sample) $\uparrow$ & $0.7519 \pm 0.0089$ & $0.7419 \pm 0.0903$ & $0.6231 \pm 0.0144$ \\
MCD $\downarrow$ & $2.7928 \pm 0.1056$ & $2.8830 \pm 0.4053$ & $3.2948 \pm 0.1236$ \\
DTW-PCC (per-sample) $\uparrow$ & $0.8184 \pm 0.0064$ & $0.8326 \pm 0.0352$ & $0.7928 \pm 0.0143$ \\
F0 RMSE (Hz) $\downarrow$ & $25.4827 \pm 1.1614$ & $28.7965 \pm 2.5670$ & $29.9511 \pm 0.6502$ \\
F0 Corr $\uparrow$ & $0.9438 \pm 0.0063$ & $0.9271 \pm 0.0120$ & $0.9204 \pm 0.0048$ \\
\bottomrule
\end{tabular}
\caption{Baseline performance across speech modes (mean $\pm$ std). Higher is better for PCC/DTW-PCC/F0 Corr, lower is better for MCD/F0 RMSE.}
\label{tab:baseline_metrics}
\end{table}

\section{Mechanistic Understanding as a Foundation for Neural Speech BCIs}

\subsection{Baseline Decoding Performance and Full Activation Patching}
\label{baseline2}

In the baseline setting, decoding performance follows a clear hierarchy across speech modes: vocalized speech is decoded accurately (PCC = 0.752, MCD = 2.8), Imagined speech achieves intermediate performance (PCC = 0.725, MCD = 2.88), and Mimed speech is decoded poorly (PCC = 0.5686, MCD = 3.34).

\begin{table}[h]
\centering

\begin{tabular}{lcc}
\toprule
\textbf{Condition} & \textbf{PCC} $\uparrow$ & \textbf{MCD} $\downarrow$ \\
\midrule
\multicolumn{3}{l}{\textit{Baseline decoding}} \\
Vocalized & 0.752 & 2.8 \\
Mimed & 0.5686 & 3.34 \\
Imagined & 0.725 & 2.88 \\
\midrule
\multicolumn{3}{l}{\textit{Full activation patching}} \\
Vocalized $\rightarrow$ Imagined & \textbf{0.954} & \textbf{1.63} \\
Vocalized $\rightarrow$ Mimed & \textbf{0.954} & \textbf{1.63} \\
Mimed $\rightarrow$ Imagined & \textbf{0.534} & \textbf{3.31} \\
Imagined $\rightarrow$ Vocalized & \textbf{0.177} & \textbf{3.70} \\
\bottomrule
\end{tabular}
\caption{Baseline decoding performance and full activation patching results. Full patching replaces either the convolutional or recurrent activations of the input mode with those from the donor mode.}
\label{tab:baseline_fullpatch}
\end{table}

Full activation patching reveals that this hierarchy is entirely determined by internal model representations. Replacing the internal activations of imagined or mimed inputs with those from vocalized speech restores and increases vocalized-level performance. Conversely, forcing vocalized inputs to use imagined-speech activations causes performance to collapse to imagined-level and further decreases the PCC. This exact numerical symmetry establishes both causal sufficiency and necessity: high-quality decoding depends on maintaining vocalized internal representations, and substituting them is sufficient to induce or destroy decoding accuracy irrespective of the input EEG signal.

These binary patching results as shown in Table~\ref{tab:baseline_fullpatch}, motivate finer-grained analyses of how causal information is distributed and interpolated within internal representations.

\section{Tri-Modal Activation Interpolation}
\label{tri-modal}

\begin{figure}[h]
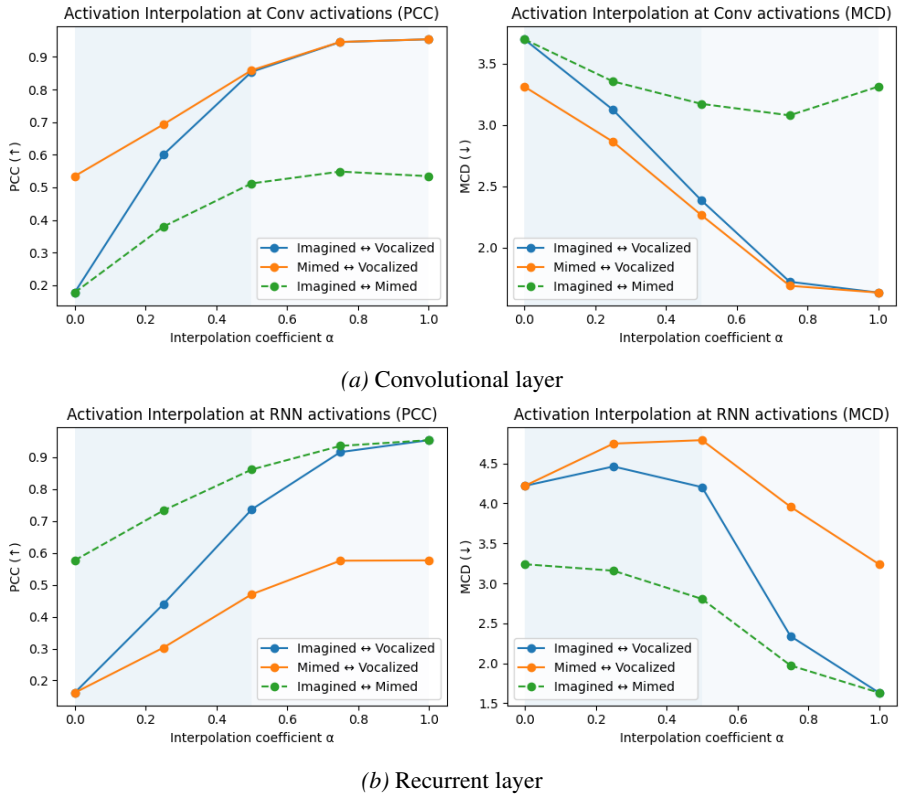

    \centering
    \begin{subfigure}[t]{\linewidth}
        \centering
        \includegraphics[width=0.7\linewidth]{figures/trimodal_interpolation_conv.pdf}
        \caption{Convolutional layer}
    \end{subfigure}
    \vspace{0.5em}
    \begin{subfigure}[t]{\linewidth}
        \centering
        \includegraphics[width=0.7\linewidth]{figures/trimodal_interpolation_rnn.pdf}
        \caption{Recurrent layer}
    \end{subfigure}
    \caption{Tri-modal activation interpolation across model layers. Decoding performance varies smoothly with the interpolation coefficient $\alpha$, with mimed speech consistently occupying an intermediate representational position.}
    \label{fig:trimodal_interp}
\end{figure}

Across all tri-modal pairs, activation interpolation produces smooth and monotonic changes in decoding quality, with symmetric behavior across interpolation directions. This indicates that imagined, mimed, and vocalized speech are not represented as discrete categories, but instead lie along a shared continuous causal axis within the model (Figure~\ref{fig:trimodal_interp}).

Interpolation effects differ across layers. At the convolutional layer, both PCC and MCD vary in a near-linear fashion with the interpolation coefficient, indicating smoothly compositional encoding of local articulatory and spectral structure. In contrast, interpolation at the recurrent layer exhibits less stable intermediate behavior, suggesting that recurrent representations encode higher-level temporal organization that requires stronger alignment to achieve consistent decoding improvements.

\section{Causal Localization and Subspace Concentration}
\label{causal_loc}

Targeted activation patching localizes cross-mode causal information to specific internal regions  as shown in Table~\ref{tab:causal_localization_horizontal}. Coarse grouping and sliding-window interventions reveal a narrow temporal bottleneck in the recurrent states and a small subset of convolutional feature channels that dominate transfer effects. Together, these results indicate that cross-modal speech decoding relies on structured, low-dimensional internal subspaces rather than uniformly distributed representations.

\begin{table*}[h]
\centering
\begin{tabular}{lccccccc}
\toprule
\textbf{Condition}
& \textbf{Conv g0} 
& \textbf{Conv g1} 
& \textbf{Conv g2} 
& \textbf{Conv g3} 
& \textbf{RNN Early} 
& \textbf{RNN Mid} 
& \textbf{RNN Late} \\
\midrule
\multicolumn{8}{l}{\textit{Sufficiency: Imagined $\leftarrow$ Vocalized}} \\
PCC 
& 0.576 & 0.535 & \textbf{0.659} & 0.503 & \textbf{0.450} & 0.427 & 0.402 \\
MCD 
& 3.23 & 3.16 & \textbf{3.00} & 3.21 & 3.29 & 3.37 & 3.41 \\
\midrule
\multicolumn{8}{l}{\textit{Necessity: Vocalized $\leftarrow$ Imagined}} \\
PCC 
& 0.940 & 0.935 & 0.932 & 0.934 & \textbf{0.698} & 0.694 & 0.686 \\
MCD 
& 1.76 & 1.79 & 1.82 & 1.83 & \textbf{2.30} & 2.41 & 2.82 \\
\bottomrule
\end{tabular}
\caption{Coarse-to-fine causal localization results reformatted horizontally. We report decoding performance when selectively patching convolutional channel groups or RNN temporal windows. Sufficiency corresponds to Imagined $\leftarrow$ Vocalized patching; necessity corresponds to Vocalized $\leftarrow$ Imagined patching.}
\label{tab:causal_localization_horizontal}
\end{table*}


\section{Modality-dependent asymmetries in neuron-level saturation analysis}
\label{app:modality_asymmetry}
The magnitude and direction of performance changes also depend strongly on the source modality. When the source modality is vocalized, i.e., when activations from the vocalized model are patched into mimed or imagined targets (green curves in Figure~\ref{fig:sat_curve}), we observe the largest improvements in PCC. This effect is related to vocalized brain responses being more strongly correlated with the acoustic structure of speech, as vocalization engages both motor activity and auditory feedback. In contrast, when the target modality is vocalized, patching from mimed or imagined activations leads to little or no improvement in PCC (Figure~\ref{fig:sat_curve}, leftmost panel), suggesting that the decoder is already operating near its optimal regime for vocalized speech.

Another interesting pattern emerges when examining interactions between mimed and imagined modalities. Imagined $\rightarrow$ mimed interventions (Figure~\ref{fig:sat_curve}, middle panel) show an initial increase in PCC followed by degradation, particularly in the convolutional layer. This improvement reflects shared neural processes between miming and imagination, potentially related to internal mechanisms such as articulatory intent or speech planning. In contrast, mimed $\rightarrow$ imagined interventions (Figure~\ref{fig:sat_curve}, rightmost panel) result in only minimal PCC improvements, indicating an asymmetry in cross-modal activation transfer between these two modalities.

\section{Sentence-level analysis of neuron patching}
\label{app:coverage_appendix}

\begin{figure*}[h]
    \centering
    \includegraphics[width=\textwidth]{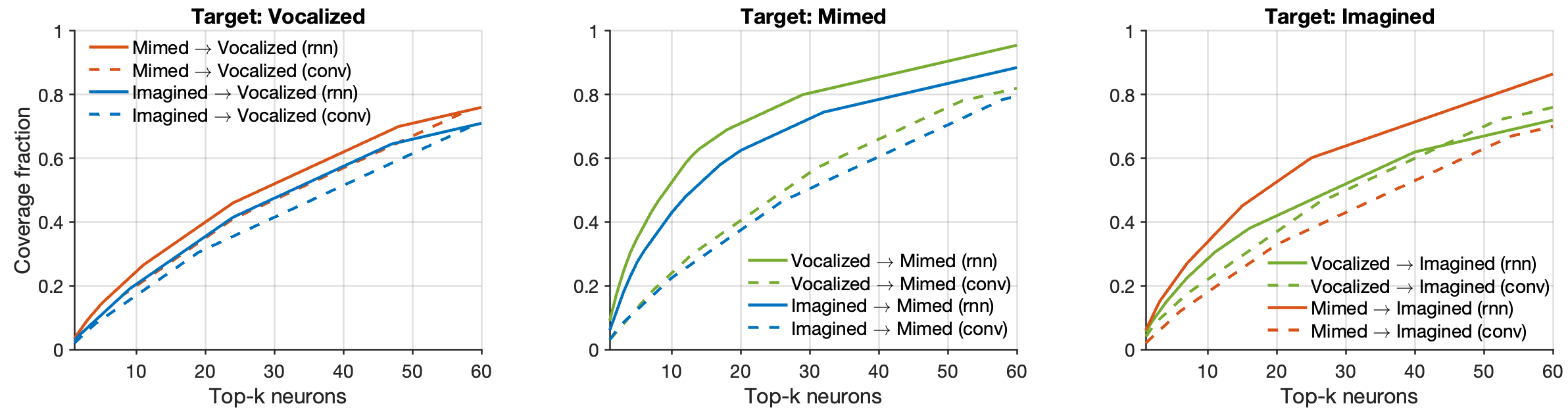}
    \caption{Sentence-level winner coverage curves for cross-modal activation patching are shown for three target modalities: Vocalized (left), Mimed (middle), and Imagined (right). Each curve reflects the fraction of sentences for which the correct target ranks within the top-k neurons after patching from a source modality.}
    \label{fig:coverage}
\end{figure*}

Figure~\ref{fig:coverage} shows sentence-level winner coverage curves for cross-modal activation patching. Across all targets (patched-to) modalities, coverage increases as a function of $k$, indicating that different sentences rely on different neurons rather than a single small set of neurons.

The rate at which the coverage curves rise provides insight into how localized or distributed the causal contributions are. Faster rises in Figure~\ref{fig:coverage} indicate more localized behavior, where a small number of neurons account for PCC improvements across many sentences, whereas slower and more gradual rises indicate more distributed neuron involvement.

A first observation is that when the target modality is mimed, the coverage curves exhibit the steepest rise. This suggests that influential neurons are more consistent across sentences when the target is mimed, especially when patching from the vocalized modality. This pattern is consistent with the saturation curves in Figure~\ref{fig:sat_curve}, where mimed targets showed the largest PCC improvements when patched from vocalized. From a neuroscience perspective, mimed speech shares motor structure with vocalized speech, and the decoder appears to reuse a common articulatory subspace, which manifests here as higher sentence-level coverage.

Another observation relates to model architecture. Neurons in the convolutional layer (dashed lines in Figure~\ref{fig:coverage}) consistently show lower coverage at all values of $k$ compared to neurons in the recurrent layer. This indicates that influential neurons are more distributed in the convolutional layer, whereas the recurrent layer relies on a smaller set of neurons that generalize across sentences. This suggests that the recurrent layer encodes more sentence-level abstractions, while the convolutional layer captures more local and sentence-specific information.

Finally, the shape of the coverage curves depends on the source–target modality pair. When the source modality is vocalized (green curves in Figure~\ref{fig:coverage}), coverage increases more rapidly. In particular, in the Vocalized $\rightarrow$ Mimed condition, the curve shows the steepest rise, with the top-5 neurons covering approximately 35\% of the sentence set. This suggests that vocalized representations activate more consistent neuron subsets when transferred to other modalities. In contrast, when the target modality is vocalized, coverage grows more slowly and almost linearly, indicating that PCC improvements arise from a larger and more distributed set of neurons.

A quantitative summary of sentence-level winner consistency across all source–target modality pairs and layers is reported in Table~\ref{tab:winner_consistency}.

\begin{table*}[h]
\centering
\small
\begin{tabular}{llccccc}
\toprule
Source $\rightarrow$ Target & Layer & \#Sent & \#Unique & Top-1 (\%) & Top-5 (\%) & Entropy \\
\midrule
Imagined $\rightarrow$ Mimed      & conv & 200 & 101 & 3.00  & 12.50 & 6.45 \\
                                  & rnn  & 200 &  83 & 6.00  & 27.50 & 5.77 \\
\midrule
Imagined $\rightarrow$ Vocalized  & conv & 200 & 118 & 2.00  &  9.50 & 6.73 \\
                                  & rnn  & 200 & 118 & 3.00  & 11.00 & 6.66 \\
\midrule
Mimed $\rightarrow$ Imagined      & conv & 200 & 120 & 2.00  & 10.00 & 6.73 \\
                                  & rnn  & 200 & 129 & 4.00  & 14.50 & 6.73 \\
\midrule
Mimed $\rightarrow$ Vocalized     & conv & 200 & 108 & 2.50  & 11.00 & 6.57 \\
                                  & rnn  & 200 & 108 & 3.50  & 14.50 & 6.49 \\
\midrule
Vocalized $\rightarrow$ Imagined  & conv & 200 & 108 & 4.00  & 13.50 & 6.52 \\
                                  & rnn  & 200 & 116 & 5.50  & 17.50 & 6.51 \\
\midrule
Vocalized $\rightarrow$ Mimed     & conv & 200 &  96 & 3.00  & 13.00 & 6.34 \\
                                  & rnn  & 200 &  69 & \textbf{9.00}  & \textbf{35.00} & \textbf{5.35} \\
\bottomrule
\end{tabular}
\caption{Sentence-level winner consistency statistics for single-neuron activation patching.
We report the number of unique winner neurons, Top-1 and Top-5 coverage (fraction of sentences covered by the most frequent winner neurons), and entropy of the winner distribution.
All results are computed over 200 sentences per condition.}
\label{tab:winner_consistency}
\end{table*}

\section{Why Interpretability Matters for Neuroscience-Oriented Models}
Deep neural networks are inspired by how the brain works and are built to deal with complex tasks in a way that is somewhat similar to brain processes. Although these models are not meant to exactly copy biological mechanisms, they are useful for testing if some computational structures or patterns match what we see in neural data.

Interpretability approaches, especially mechanistic interpretability, help us see how similar a trained neural network is to the brain beyond just its performance. Instead of viewing the model as a black box, these methods look at how its internal patterns are organized, how information moves through it, and which parts are responsible for important behaviors. In neuroscience, this lets us compare the model’s inner workings to known features of neural systems, like distributed patterns, bottlenecks, and layers of organization.

Seen this way, interpretability experiments act as a way to question the model. They let us find out if a network solves tasks using strategies like those in the brain or if it uses different shortcuts that still get the right answers. This difference matters, especially when models are trained on neural data, because high accuracy alone does not guarantee that the model’s patterns accurately reflect how the brain works.

There is a practical reason for this work as well. Biological systems are very robust and can adapt to different situations and noise levels. If certain features help with this robustness, then models with more brain-like organization might also adapt better, especially when there is little data or changing conditions. Interpretability techniques help test this idea by connecting a model’s inner structure to how well it generalizes, instead of just looking at its input and output.

In neuroscience, it gives researchers tools to test ideas about computation in systems they can control and change. In machine learning, interpretability helps identify which internal structures are useful for modeling neural data, allowing neural data to directly shape how models are built.

\end{document}